\documentclass[runningheads]{llncs}
\usepackage{graphicx}
\usepackage[utf8]{inputenc}
\usepackage{epsfig}
\usepackage{calc}
\usepackage{amssymb}
\usepackage{amstext}
\usepackage{amsmath}
\usepackage{multicol}
\usepackage{pslatex}
\usepackage{amsmath}
\usepackage{listings}
\usepackage[ruled,vlined,linesnumbered]{algorithm2e}
\usepackage{comment}
\usepackage{amssymb}
\usepackage{url}
\usepackage{enumitem}

\begin{document}

\title{Pushing the Envelope: From Discrete to Continuous Movements in Multi-Agent Path Finding via Lazy Encodings}
\titlerunning{From Discrete to Continuous Movements in Multi-Agent Path Finding}

%
%
\author{Pavel Surynek\orcidID{0000-0001-7200-0542}}
%
\authorrunning{P. Surynek}

%
\institute{
Faculty of Information Technology\\
Czech Technical University in Prague\\
Th\'{a}kurova 9, 160 00 Praha 6, Czechia\\
\email{pavel.surynek@fit.cvut.cz}
}


%
\maketitle              

\begin{abstract}
Multi-agent path finding in continuous space and time with geometric agents MAPF$^\mathcal{R}$ is addressed in this paper. The task is to navigate agents that move smoothly between predefined positions to their individual goals so that they do not collide. We introduce a novel solving approach for obtaining makespan optimal solutions called SMT-CBS$^\mathcal{R}$ based on {\em satisfiability modulo theories} (SMT). The new algorithm combines collision resolution known from conflict-based search (CBS) with previous generation of incomplete SAT encodings on top of a novel scheme for selecting decision variables in a potentially uncountable search space. We experimentally compare SMT-CBS$^\mathcal{R}$ and previous CCBS algorithm for MAPF$^\mathcal{R}$.

\keywords{path finding, multiple agents, robotic agents, logic reasoning, satisfiability modulo theory, makespan optimality}
\end{abstract}

\section{Introduction}

In {\em multi-agent path finding} (MAPF) \cite{DBLP:conf/focs/KornhauserMS84,DBLP:conf/aiide/Silver05,DBLP:journals/jair/Ryan08,DBLP:conf/icra/Surynek09,DBLP:journals/jair/WangB11,DBLP:journals/ai/SharonSGF13,SharonSFS15,BoteaS15} the task is to navigate agents from given starting positions to given individual goals. The problem takes place in undirected graph $G=(V,E)$ where agents from set $A=\{a_1,a_2,...,a_k\}$ are placed in vertices with at most one agent per vertex. The initial configuration can be written as $\alpha_0: A \rightarrow V$ and similarly the goal configuration as $\alpha_+: A \rightarrow V$. The task of navigating agents can be then expressed formally as transforming $\alpha_0$ into $\alpha_+$ while movements are instantaneous and are possible across edges assuming no other agent is entering the same target vertex in the standard MAPF. 

In order to reflect various aspects of real-life applications, variants of MAPF have been introduced such as those considering {\em kinematic constraints} \cite{DBLP:conf/ijcai/HonigK00XAK17}, {\em large agents} \cite{LargeAAAI2019}, {\em generalized costs} of actions \cite{DBLP:conf/ijcai/WalkerSF18}, or {\em deadlines} \cite{DBLP:conf/ijcai/0001WFLKK18} - see \cite{DBLP:journals/corr/0001KA0HKUXTS17,DBLP:conf/raai/Stern19} for more variants. Particularly in this work we are dealing with an extension of MAPF introduced only recently \cite{DBLP:conf/ijcai/AndreychukYAS19,DBLP:conf/icaart/Surynek20} that considers continuous time and space (MAPF$^\mathcal{R}$) where agents move smoothly along predefined curves interconnecting predefined positions placed arbitrarily in some continuous space. It is natural in MAPF$^\mathcal{R}$ to assume geometric agents of various shapes that occupy certain volume in the space -  circles in the 2D space, polygons, spheres in the 3D space etc. In contrast to MAPF, where the collision is defined as the simultaneous occupation of a vertex or an edge by two agents, collisions are defined as any spatial overlap of agents' bodies in MAPF$^\mathcal{R}$. 

The motivation behind introducing MAPF$^\mathcal{R}$ is the need to construct more realistic paths in many applications such as controlling fleets of robots or aerial drones \cite{DBLP:conf/atal/JanovskyCV14,DBLP:conf/atal/CapNVP13} where continuous reasoning is closer to the reality than the standard MAPF.

We contribute by showing how to apply satisfiability modulo theory (SMT) reasoning \cite{DBLP:journals/constraints/BofillPSV12,DBLP:conf/cp/Nieuwenhuis10} in makespan optimal MAPF$^\mathcal{R}$ solving. Particularly we extend the preliminary work in this direction from \cite{DBLP:journals/corr/abs-1903-09820,DBLP:conf/socs/Surynek19,DBLP:conf/icaart/Surynek20}. The SMT paradigm constructs decision procedures for various complex logic theories by decomposing the decision problem into the propositional part having arbitrary Boolean structure and the complex theory part that is restricted on the conjunctive fragment. Our SMT-based algorithm called SMT-CBS$^\mathcal{R}$ combines the Conflict-based Search (CBS) algorithm \cite{SharonSFS15,DBLP:conf/aips/FelnerLB00KK18} with previous algorithms for solving the standard MAPF using incomplete encodings \cite{DBLP:conf/ijcai/Surynek19,DBLP:conf/iros/Surynek19,DBLP:conf/icaart/Surynek19} and continuous reasoning.

\subsection{Related Work and Organization}

Using reductions of planning problems to propositional satisfiability has been coined in the SATPlan algorithm and its variants \cite{DBLP:conf/ecai/KautzS92,DBLP:conf/aaai/KautzS96,DBLP:conf/ijcai/KautzS99,DBLP:conf/aaai/Kautz06}. Here we are trying to apply similar idea in the context of MAPF$^\mathcal{R}$. So far MAPF$^\mathcal{R}$ has been solved by a modified version of CBS that tries to solve MAPF lazily by adding collision avoidance constraints on demand. The adaptation of CBS for MAPF$^\mathcal{R}$ consists in implementing continuous collision detection while the high-level framework of the algorithm remains the same as demonstrated in the CCBS algorithm \cite{DBLP:conf/ijcai/AndreychukYAS19}.

We follow the idea of CBS too but instead of searching the tree of possible collision eliminations at the high-level we encode the requirement of having collision free paths as a propositional formula \cite{Biere:2009:HSV:1550723} and leave it to the SAT solver as done in \cite{SurynekFSB16}. We construct the formula {\em lazily} by adding collision elimination refinements following \cite{DBLP:conf/ijcai/Surynek19} where the lazy construction of incomplete encodings has been suggested for the standard MAPF within the algorithm called SMT-CBS. SMT-CBS works with propositional variables indexed by {\em agent} $a$, {\em vertex} $v$, and {\em time step} $t$ with the meaning that if the variable is $\mathit{TRUE}$ $a$ in $v$ at time step $t$. In MAPF$^\mathcal{R}$ we however face major technical difficulty that we do not know necessary decision (propositional) variables in advance and due to continuous time we cannot enumerate them all as in the standard MAPF. Hence we need to select from a potentially uncountable space those variables that are sufficient for finding the makespan optimal solution.

The organization is as follows: we first introduce MAPF$^\mathcal{R}$. Then we recall CCBS, a variant of CBS for MAPF$^\mathcal{R}$. Details of the novel SMT-based solving algorithm SMT-CBS$^\mathcal{R}$ follow. Finally, a comparison SMT-CBS$^\mathcal{R}$ with CCBS is shown.

\subsection{MAPF in Continuous Time and Space}
We use the definition of MAPF in continuous time and space denoted MAPF$^\mathcal{R}$ from \cite{DBLP:conf/ijcai/WalkerSF18} and \cite{DBLP:conf/ijcai/AndreychukYAS19}. MAPF$^\mathcal{R}$ shares components with the standard MAPF: undirected graph $G=(V,E)$, set of agents $A=\{a_1,a_2,...,a_k\}$, and the initial and goal configuration of agents:  $\alpha_0: A \rightarrow V$ and $\alpha_+: A \rightarrow V$. A simple 2D variant of MAPF$^\mathcal{R}$ is as follows:

\begin{definition} ({\bf MAPF$^\mathcal{R}$}) Multi-agent path finding with continuous time and space (MAPF$^\mathcal{R}$) is a 5-tuple $\Sigma^\mathcal{R}=(G=(V,E), A, \alpha_0, \alpha_+, \rho)$ where $G$, $A$, $\alpha_0$, $\alpha_+$ are from the standard MAPF and $\rho$ determines continuous extensions:
\begin{itemize}
	\item $\rho.x(v), \rho.y(v)$ for $v \in V$ represent the position of vertex $v$ in the 2D plane
	\item $\rho.speed(a)$ for $a \in A$ determines constant speed of agent $a$
	\item $\rho.radius(a)$ for $a \in A$ determines the radius of agent $a$; we assume that agents are circular discs with omni-directional ability of movements
\end{itemize}
\end{definition}

We assume that agents have constant speed and instant acceleration. The major difference from the standard MAPF where agents move instantly between vertices (disappears in the source and appears in the target instantly) is that smooth continuous movement between a pair of vertices (positions) along the straight line interconnecting them takes place in MAPF$^\mathcal{R}$. Hence we need to be aware of the presence of agents at some point in the 2D plane at any time.

\begin{figure}[h]
    \centering
    \vspace{-0.1cm}
    \includegraphics[trim={2.5cm 15.1cm 8.0cm 2.5cm},clip,width=0.65\textwidth]{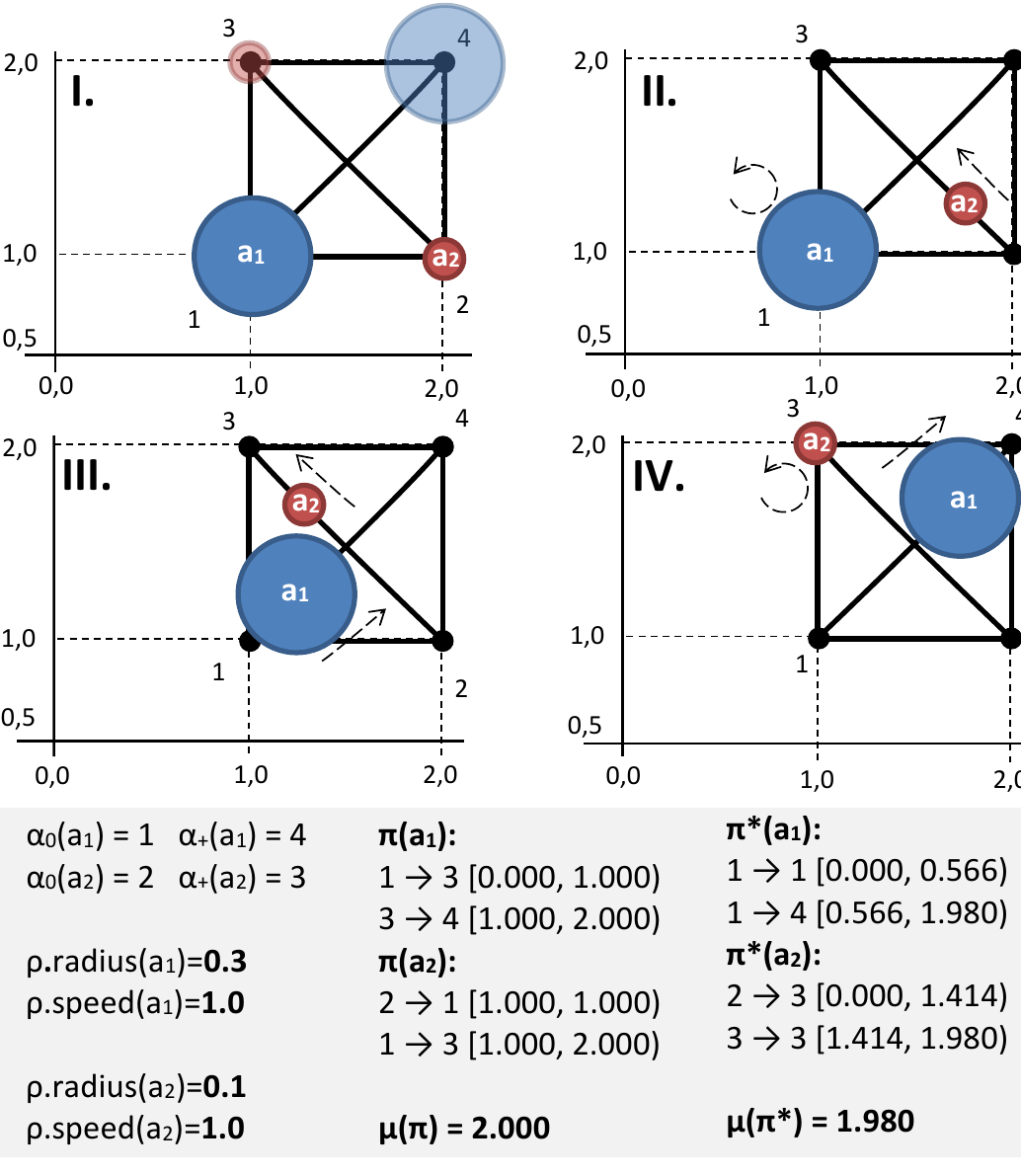}
    \vspace{-0.4cm}
    \caption{An example of MAPF$^\mathcal{R}$ instance with two agents. A feasible makespan sub-optimal solution $\pi$ (makespan $\mu(\pi)=2.0$) and makespan optimal solution $\pi*$ (makespan $\mu(\pi*)=1.980$) are shown.}
    \label{fig-MAPF-R_nuovo}
\end{figure}

Collisions may occur between agents in MAPF$^\mathcal{R}$ due to their volume; that is, they collide whenever their bodies {\bf overlap}. In contrast to MAPF, collisions in MAPF$^\mathcal{R}$ may occur not only in a single vertex or edge being shared by colliding agents but also on pairs of edges (lines interconnecting vertices) that are too close to each other and simultaneously traversed by large agents.

We can further extend the continuous properties by introducing the direction of agents and the need to rotate agents towards the target vertex before they start to move. Also agents can be of various shapes not only circular discs \cite{LargeAAAI2019} and can move along various fixed curves.

For simplicity we elaborate our implementations for the above simple 2D continuous extension with circular agents. We however note that all developed concepts can be adapted for MAPF with more continuous extensions.

A solution to given MAPF$^\mathcal{R}$ $\Sigma^\mathcal{R}$ is a collection of temporal plans for individual agents $\pi = [\pi(a_1),$ $\pi(a_2),...,$ $\pi(a_k)]$ that are {\bf mutually collision-free}. A temporal plan for agent $a \in A$ is a sequence $\pi(a) = [((\alpha_0(a),\alpha_1(a)),$ $[t_0(a),t_1(a)));$ $((\alpha_1(a),\alpha_2(a)),$ $[t_1(a),t_2(a)));$ ...; $((\alpha_{m(a)-1},\alpha_{m(a)}(a)),$ $[t_{m(a)-1},t_{m(a)}))]$ where $m(a)$ is the length of individual temporal plan and each pair $(\alpha_i(a),\alpha_{i+1}(a)),$ $[t_i(a),t_{i+1}(a)))$ corresponds to traversal event between a pair of vertices $\alpha_i(a)$ and $\alpha_{i+1}(a)$ starting at time $t_i(a)$ and finished at $t_{i+1}(a)$.

It holds that $t_i(a) < t_{i+1}(a)$ for $i = 0,1,...,m(a)-1$. Moreover consecutive events in the individual temporal plan must correspond to edge traversals or waiting actions, that is: $\{\alpha_i(a),$ $\alpha_{i+1}(a)\} \in E$ or $\alpha_i(a) = \alpha_{i+1}(a)$; and times must reflect the speed of agents for non-wait actions.

 
The duration of individual temporal plan $\pi(a)$ is called an {\em individual makespan}; denoted $\mu(\pi(a)) = t_{m(a)}$. The overall {\em makespan} of $\pi$ is defined as max$_{i=1}^k(\mu(\pi(a_i)))$. We focus on finding makespan optimal solutions. An example of MAPF$^\mathcal{R}$ and makespan optimal solution is shown in Figure \ref{fig-MAPF-R_nuovo}.

Through straightforward reduction of MAPF to MAPF$^\mathcal{R}$ it can be observed that finding a makespan optimal solution with continuous time is an NP-hard problem \cite{DBLP:journals/jsc/RatnerW90,DBLP:journals/corr/YuL15c}.

\section{Solving MAPF with Continuous Time}

Let us recall CCBS \cite{DBLP:conf/ijcai/AndreychukYAS19}, a variant of CBS \cite{SharonSFS15} modified for MAPF$^\mathcal{R}$. The idea of CBS algorithms is to resolve conflicts lazily. CBS algorithms are usually developed for the {\em sum-of-costs} \cite{DBLP:journals/ai/SharonSGF13} objective but using other cumulative costs like makespan is possible too.

\subsection{Conflict-based Search}

CCBS for finding the makespan optimal solution is shown in Algorithm \ref{alg-CCBS}. The high-level of CCBS searches a {\em constraint tree} (CT) using a priority queue ordered according to the makespan in the breadth first manner. CT is a binary tree where each node $N$ contains a set of collision avoidance constraints $N.cons$ - a set of triples $(a_i,(u,v),[t_0,t_+))$ forbidding agent $a_i$ to start smooth traversal of edge $\{u,v\}$ (line) at any time between $[t_0,t_+)$, a solution $N.\pi$ - a set of $k$ individual temporal plans, and the makespan $N.\mu$ of $N.\pi$.

\begin{algorithm}[t]
\begin{footnotesize}
\SetKwBlock{NRICL}{CBS$^\mathcal{R}$ ($\Sigma^\mathcal{R}=(G=(V,E), A, \alpha_0, \alpha_+, \rho)$)}{end} \NRICL{
    $R.cons \gets \emptyset$ \\
    $R.\pi \gets$ $\{$shortest temporal plan from $\alpha_0(a_i)$ to $\alpha_+(a_i)\;|\;i = 1,2,...,k\}$\\
    $R.\mu \gets \max_{i=1}^k{\mu(N.\pi(a_i))}$ \\
    $\textsc{Open} \gets \emptyset$ \\     
    insert $R$ into $\textsc{Open}$ \\
    \While {$\textsc{Open} \neq \emptyset$} {
        $N \gets$ min$_{\mu}(\textsc{Open}$)\\
        remove-Min$_{\mu}(\textsc{Open}$)\\
        $collisions \gets$ validate-Plans($N.\pi$) \\
        \If {$collisions = \emptyset$}{
            \Return $N.\pi$\\
        }
        {\bf let} $(m_i \times m_j) \in collisions$ where $m_i = (a_i,(u_i,v_i),[t_i^0,t_i^+))$ and $m_j=(a_j,(u_j,v_j),[t_j^0,t_j^+))$\\
        $([\tau_i^0, \tau_i^+);[\tau_j^0, \tau_j^+)) \gets$ resolve-Collision($m_i$,$m_j$) \\
        
        \For {each $m \in \{(m_i,[\tau_i^0, \tau_i^+)),(m_j,[\tau_j^0, \tau_j^+))\}$}{
             let $((a,(u,v),[t_0,t_+)),[\tau_0, \tau_+))=m$\\
       	$N'.cons \gets N.cons\cup \{(a,(u,v),[\tau_0,\tau_+))\}$\\
        	$N'.\pi \gets N.\pi$\\
        	update($a$, $N'.\pi$, $N'.cons$)\\
		$N'.\mu \gets \max_{i=1}^k\mu{(N'.\pi(a_i))}$\\
		insert $N'$ into $\textsc{Open}$ \\
        }
     }
} \caption{CCBS algorithm for solving MAPF$^\mathcal{R}$.} \label{alg-CCBS}
\end{footnotesize}
\end{algorithm}

The low-level in CCBS associated with node $N$ searches for individual temporal plan with respect to set of constraints $N.cons$. For given agent $a_i$, this is the standard single source shortest path search from $\alpha_0(a_i)$ to $\alpha_+(a_i)$ that at time $t$ cannot start to traverse any $\{(u,v) \in E\;|\;(a_i,(u,v),[t_0,t_+)) \in N.cons \wedge t \in [t_0,t_+)\}$. Various intelligent single source shortest path algorithms such as SIPP \cite{DBLP:conf/icra/PhillipsL11} can be used here.

CCBS stores nodes of CT into priority queue $\textsc{Open}$ sorted according to the ascending makespan. At each step CBS takes node $N$ with the lowest makespan from $\textsc{Open}$ and checks if $N.\pi$ represents non-colliding temporal plans. If there is no collision, the algorithms returns valid solution $N.\pi$. Otherwise the search branches by creating a new pair of nodes in CT - successors of $N$. Assume that a collision occurred between $a_i$ traversing $(u_i,v_i)$ during $[t_i^0,t_i^+)$ and $a_j$ traversing $(u_j,v_j)$ during $[t_j^0,t_j^+)$. This collision can be avoided if either agent $a_i$ or agent $a_j$ waits after the other agent passes. We can calculate for $a_i$ so called maximum {\em unsafe interval} $[\tau_i^0,\tau_i^+)$ such that whenever $a_i$ starts to traverse $(u_i,v_i)$ at some time $t \in [\tau_i^0,\tau_i^+)$ it ends up colliding with $a_j$ assuming $a_j$ did not try to avoid the collision. Hence $a_i$ should wait until $\tau_i^+$ to tightly avoid the collision with $a_j$. Similarly we can calculate maximum unsafe interval for $a_j$: $[\tau_j^0,\tau_j^+)$. These two options correspond to new successor nodes of $N$: $N_1$ and $N_2$ that inherit set of constraints from $N$ as follows: $N_1.cons = N.cons$ $\cup$ $\{(a_i,(u_i,v_i),$ $[\tau_i^0, \tau_i^+))\}$ and $N_2.cons = N.cons$ $\cup$ $\{(a_j,(u_j,v_j),$ $[\tau_j^0, \tau_j^+))\}$. $N_1.\pi$ and $N_1.\pi$ inherits plans from $N.\pi$ except those for agents $a_i$ and $a_j$ respectively that are recalculated with respect to the constraints. After this $N_1$ and $N_2$ are inserted into $\textsc{Open}$.

\subsection{A Satisfiability Modulo Theory Approach}

We will use for the specific case of CCBS the idea introduced in \cite{DBLP:conf/ijcai/Surynek19} that rephrases the algorithm as problem solving in {\em satisfiability modulo theories} (SMT) \cite{DBLP:journals/constraints/BofillPSV12,DBLP:conf/wollic/Tinelli10}. The basic use of SMT divides the satisfiability problem in some complex theory $T$ into a propositional part that keeps the Boolean structure of the problem and a simplified procedure $\mathit{DECIDE_T}$ that decides fragment of $T$ restricted on {\em conjunctive formulae}. A general $T$-formula $\Gamma$ being decided for satisfiability is transformed to a {\em propositional skeleton} by replacing its atoms with propositional variables. The standard SAT solver then decides what variables should be assigned $\mathit{TRUE}$ in order to satisfy the skeleton - these variables tells what atoms hold in $\Gamma$. $\mathit{DECIDE_T}$ then checks if the conjunction of atoms assigned $\mathit{TRUE}$ is valid with respect to axioms of $T$. If so then satisfying assignment is returned. Otherwise a conflict from $\mathit{DECIDE_T}$ (often called a {\em lemma}) is reported back to the SAT solver and the skeleton is extended with new constraints resolving the conflict. More generally not only new constraints are added to resolve the conflict but also new atoms can be added to $\Gamma$.


$T$ will be represented by a theory with axioms describing movement rules of MAPF$^\mathcal{R}$; a theory we will denote $T_{\mathit{MAPF}^\mathcal{R}}$. $\mathit{DECIDE}_{\mathit{MAPF}^\mathcal{R}}$ can be naturally represented by the plan validation procedure from CCBS (validate-Plans).



\subsection{RDD: Real Decision Diagram}

The important question when using the logic approach is what will be the decision variables. In the standard MAPF, time expansion of $G$ for every time step has been done resulting in a multi-value decision diagram (MDD) \cite{SurynekFSB16} representing possible positions of agents at any time step. Since MAPF$^\mathcal{R}$ is inherently continuous we cannot afford to consider every time moment but we need to restrict on important moments only.

Analogously to MDD, we introduce {\em real decision diagram} (RDD). RDD$_i$ defines for agent $a_i$ its space-time positions and possible movements. Formally, $RDD_i$ is a directed graph $(X^i,E^i)$ where $X_i$ consists of pairs $(u,t)$ with $u \in V$ and $t \in \mathbb{R}_0^+$ is time and $E_i$ consists of directed edges of the form $((u,t_u);(v,t_v))$. Edge $((u,t_u);(v,t_v))$ correspond to agent's movement from $u$ to $v$ started at $t_u$ and finished at $t_v$. Waiting in $u$ is possible by introducing edge $((u,t_u);(v,t'_u))$. Pair $(\alpha_0(a_i),0) \in X_i$ indicates start and $(\alpha_+(a_i),t)$ for some $t$ corresponds to reaching the goal position.

\begin{algorithm}[t]
\begin{footnotesize}
\SetKwBlock{NRICL}{build-RDDs($\Sigma^\mathcal{R}$, $\mathit{cons}$, $\mu_{max}$)}{end} \NRICL{
    \For {$i = 1,2,...,k$}{
    $X^i \gets \emptyset$, $E^i \gets \emptyset$, $\textsc{Open} \gets \emptyset$ \\   
    insert $(\alpha_0(a_i),0)$ into $\textsc{Open}$\\
    $X^i \gets X^i \cup \{ (\alpha_0(a_i), 0) \}$\\
    
    \While {$\textsc{Open} \neq \emptyset$} {
            $(u,t) \gets$ min$_{t}(\textsc{Open}$)\\
            remove-Min$_{t}(\textsc{Open}$) \\
            \If {$t \leq \mu_{max}$}{
	            \For {each $v \;|\; \{u,v\} \in E$}{
      		             $\Delta t \gets dist(u,v) / v_{a_i}$ \\
             		insert $(v,t+\Delta t)$ into $\textsc{Open}$\\             		
				$X^i \gets X^i \cup \{ (v, t+\Delta t) \}$\\
				$E^i \gets E^i \cup \{ [(u,t); (v,t+\Delta t)] \}$ \\
	    			\For {each $(a_i, (u,v), [\tau^0,\tau^+)) \in \mathit{cons}$}{
            				\If {$t \geq \tau^0$ and $t < \tau^+$}{
             				insert $(u,\tau_+)$ into $\textsc{Open}$\\
						$X^i \gets X^i \cup \{ (u, \tau^+) \}$ \\
						$E^i \gets E^i \cup \{ [(u,t);(u, \tau^+)] \}$ \\
             			}
	            		}
      		      	}
		}
	  }
    }
    \Return $[(X^1,E^1),(X^2,E^2),...,(X^k,E^k)]$\\
} \caption{Building of RDD for MAPF$^\mathcal{R}$}
\label{alg-RDD-build}
\end{footnotesize}
\end{algorithm}

RDDs for individual agents are constructed with respect to collision avoidance constraints. If there is no collision avoidance constraint then RDD$_i$ simply corresponds to a shortest temporal plan for agent $a_i$. But if a collision avoidance constraint is present, say $(a_i, (u,v), [\tau_0,\tau_+))$, and we are considering movement starting in $u$ at $t$ that interferes with the constraint, then we need to generate a node into RDD$_i$ that allows agent to wait until the unsafe interval passes by, that is node $(u,\tau^+)$ and edge $((u,\tau^+);(u,\tau^+))$ are added.

\begin{figure}[h]
    \centering
    \vspace{-0.1cm}
    \includegraphics[trim={2.6cm 19.5cm 7.5cm 3.0cm},clip,width=0.75\textwidth]{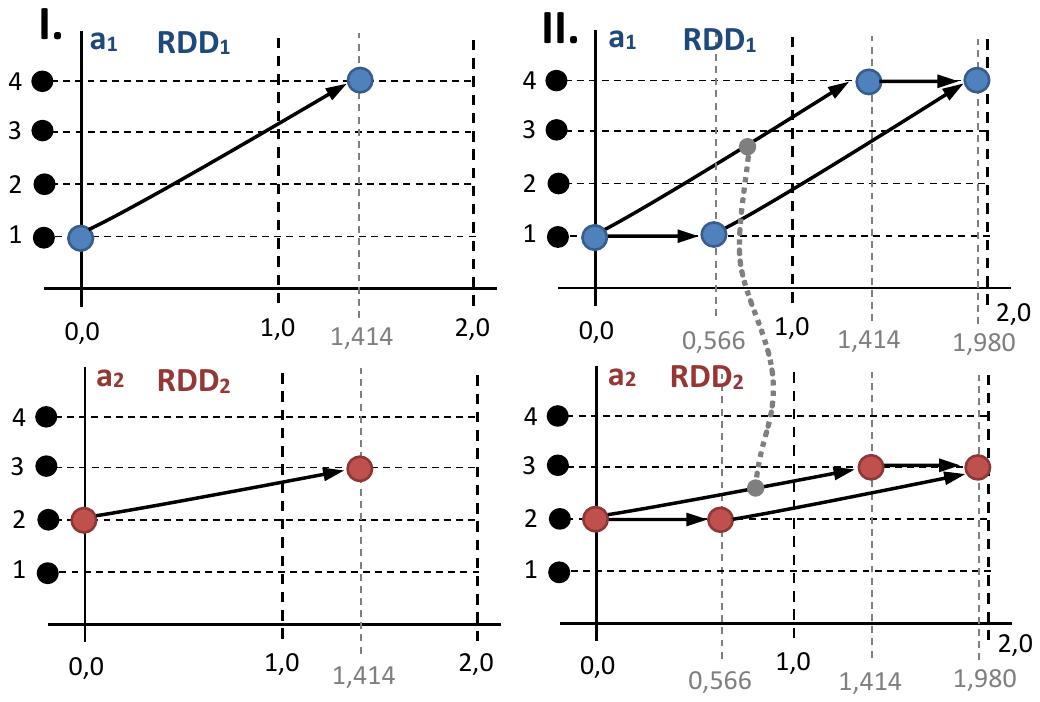}
    \vspace{-0.4cm}
    \caption{Real decision diagrams (RDDs) for agents $a_1$ and $a_2$ from MAPF$^\mathcal{R}$ from Figure \ref{fig-MAPF-R_nuovo}. Decisions corresponding to shortest paths for agents $a_1$ and $a_2$ moving diagonally towards their goals are shown: $a_1: 1 \rightarrow 4$, $a_2: 2 \rightarrow 3$ (left). This however results in a collision whose resolution is either waiting for agent $a_1$ in vertex 1 from $0.000$ until $0.566$ or waiting for agent $a_2$ in vertex 2 from $0.000$ until $0.566$; reflected in the next RDDs (right). Mutex is depicted using dotted line connecting arcs form RDD$_1$ and RDD$_2$. }
    \label{fig-RDD}
\end{figure}

The process of building RDDs is formalized in Algorithm \ref{alg-RDD-build}. It performs breadth-first search (BFS). For each possible edge traversal the algorithm generates a successor node and corresponding edge (lines 12-15) but also considers all possible wait action w.r.t. interfering collision avoidance constraints (lines 17-20). As a result each constraint is treated as both present and absent. In other words, RDD$_i$ represents union of all paths for agent $a_i$ examined in all branches of the corresponding CT. The stop condition is specified by the maximum makespan $\mu_{max}$ beyond which no more actions are performed. An example of RDDs is shown in Figure \ref{fig-RDD}.

\subsection{SAT Encoding from RDD}

We introduce a decision variable for each node and edge $[RDD{_1},...,RDD{_k}]$; RDD$_i=(X^i,E^i)$: we have variable $\mathcal{X}_{u}^{t}(a_i)$ for each $(u,t) \in X^i$ and $\mathcal{E}_{u,v}^{t_u,t_v}(a_i)$ for each directed edge $((u,t_u);(v,t_v)) \in E^i$. The meaning of variables is that $\mathcal{X}_{u}^{t}(a_i)$ is $\mathit{TRUE}$ if and only if agent $a_i$ appears in $u$ at time $t$ and similarly for edges: $\mathcal{E}_{u,v}^{t_u,t_v}(a_i)$ is $\mathit{TRUE}$ if and only if $a_i$ moves from $u$ to $v$ starting at time $t_u$ and finishing at $t_v$.

MAPF$^\mathcal{R}$ rules are encoded on top of these variables so that eventually we want to obtain formula $\mathcal{F}(\mu)$ that encodes existence of a solution of makespan $\mu$ to given MAPF$^\mathcal{R}$. We need to encode that agents do not skip but move along edges, do not disappear or appear from nowhere etc. We show below constraints stating that if agent $a_i$ appears in vertex $u$ at time step $t_u$ then it has to leave through exactly one edge connected to $u$ (constraint (\ref{eq-2}) although Pseudo-Boolean can be encoded using purely propositional means):

\begin{equation}
   {  \mathcal{X}_u^{t_u}(a_i) \Rightarrow \bigvee_{(v,t_v)\;|\;((u,t_u),(v,t_v)) \in E^i}{\mathcal{E}^{t_u,t_v}_{u,v}(a_i)},
   }
   \label{eq-1}
\end{equation}

\begin{equation}
   {  \sum_{(v,t_v)\;|\;((u,t_u),(v,t_v)) \in E^i }{\mathcal{E}_{u,v}^{t_u,t_v}{(a_i)} \leq 1}
   }
   \label{eq-2}
\end{equation}

\begin{equation}
   {  {\mathcal{E}_{u,v}^{t_u,t_v}{(a_i)} \Rightarrow \mathcal{X}_v^{t_v}(a_i) }
   }
   \label{eq-3}
\end{equation}

Analogously to (\ref{eq-2}) we have constraint allowing a vertex to accept at most one agent through incoming edges; plus we need to enforce agents starting in $\alpha_0$ and finishing in $\alpha_+$.

\begin{proposition}
Any satisfying assignment of $\mathcal{F}(\mu)$ correspond to valid individual temporal plans for $\Sigma^\mathcal{R}$ whose makespans are at most $\mu$.
\label{prop-valid}
\end{proposition}

We apriori do not add constraints for eliminating collisions; these are added lazily after assignment/solution validation. Hence, $\mathcal{F}(\mu)$ constitutes an {\em incomplete model} for $\Sigma^\mathcal{R}$: $\Sigma^\mathcal{R}$ is solvable within makespan $\mu$ then $\mathcal{F}(\mu)$ is satisfiable. The opposite implication does not hold since satisfying assignment of $\mathcal{F}(\mu)$ may lead to a collision.

From the perspective of SMT, the propositional level does not understand geometric properties of agents so cannot know what simultaneous variable assignments are invalid. This information is only available at the level of theory $T =$ MAPF$^\mathcal{R}$ through $\mathit{DECIDE}_{\mathit{MAPF}^\mathcal{R}}$.


\subsection{Lazy Encoding via Mutex Refinements}

The SMT-based algorithm itself is divided into two procedures: SMT-CBS$^\mathcal{R}$ representing the main loop and SMT-CBS-Fixed$^\mathcal{R}$ solving the input MAPF$^\mathcal{R}$ for a fixed maximum makespan $\mu$. The major difference from the standard CBS is that there is {\bf no branching} at the high-level.

Procedures {\em encode-Basic} and {\em augment-Basic} in Algorithm \ref{alg-SMTCBS-low} build formula $\mathcal{F}(\mu)$ according to given RDDs and the set of collected collision avoidance constraints. New collisions are resolved {\bf lazily} by adding {\em mutexes} (disjunctive constraints). A collision is avoided in the same way as in CCBS; that is, one of the colliding agent waits. Collision eliminations are tried until a valid solution is obtained (line 11) or until a failure for current $\mu$ (line 20) which means to try bigger makespan.

For resolving a collision we need to: {\bf (1)} eliminate simultaneous execution of colliding movements and {\bf (2)} augment the formula to enable avoidance (waiting). Assume a collision between agents $a_i$ traversing $(u_i,v_i)$ during $[t_i^0,t_i^+)$ and $a_j$ traversing $(u_j,v_j)$ during $[t_j^0,t_j^+)$ which corresponds to variables $\mathcal{E}_{u_i,v_i}^{t_i^0,t_i^+}(a_i)$ and $\mathcal{E}_{u_j,v_j}^{t_j^0,t_j^+}(a_j)$. The collision can be eliminated by adding the following {\bf mutex} (disjunction) to the formula: $\neg \mathcal{E}_{u_i,v_i}^{t_i^0,t_i^+}(a_i) \vee \neg \mathcal{E}_{u_j,v_j}^{t_j^0,t_j^+}(a_j)$ (line 13 in Algorithm \ref{alg-SMTCBS-low}). Satisfying assignments of the next $\mathcal{F}(\mu)$ can no longer lead to this collision. Next, the formula is augmented according to new RDDs that reflect the collision - decision variables and respective constraints are added.


\begin{algorithm}[h]
\begin{footnotesize}
\SetKwBlock{NRICL}{SMT-CBS$^\mathcal{R}$ ($\Sigma^\mathcal{R}=(G=(V,E), A, \alpha_0, \alpha_+, \rho)$)}{end} \NRICL{
    $\mathit{constraints} \gets \emptyset$\\
    $\pi \gets$ $\{\pi^*(a_i)$ a shortest temporal plan from $\alpha_0(a_i)$ to $\alpha_+(a_i)\;|\;i = 1,2,...,k\}$ \\
    $\mu \gets \max_{i=1}^k{\mu(\pi(a_i))}$ \\
    \While {$\mathit{TRUE}$}{
         $(\pi,constraints,\mu_{next}) \gets$\\
         $\;\;$SMT-CBS-Fixed$^\mathcal{R}$($\Sigma^\mathcal{R}$, $constraints$, $\mu$)\\
        \If {$\pi \neq$ UNSAT}{
        	\Return $\pi$\\
        }
        $\mu \gets \mu_{next}$\\
    }
}
\caption{High-level of SMT-CBS$^\mathcal{R}$} \label{alg-SMTCBS-high}
\end{footnotesize}
\end{algorithm}

The set of pairs of collision avoidance constraints is propagated across entire execution of the algorithm. Constraints originating from a single collision are grouped in pairs so that it is possible to introduce mutexes for colliding movements discovered in previous steps.

Algorithm \ref{alg-SMTCBS-high} shows the main loop of SMT-CBS$^\mathcal{R}$. The algorithm checks if there is a solution for $\Sigma^\mathcal{R}$ of makespan $\mu$. It starts at the lower bound for $\mu$ obtained as the duration of the longest from shortest individual temporal plans ignoring other agents (lines 3-4). Then $\mu$ is iteratively increased in the main loop (lines 5-9) following the style of SATPlan \cite{DBLP:conf/ijcai/KautzS99}. The algorithm relies on the fact that the solvability of MAPF$^\mathcal{R}$ w.r.t. cumulative objective like the makespan behaves as a non decreasing function. Hence trying increasing makespans eventually leads to finding the optimum provided we do not skip any relevant makespan.

The next makespan to try will then be obtained by taking the current makespan plus the smallest duration of the continuing movement (lines 17-18 of Algorithm \ref{alg-SMTCBS-low}). The following proposition is a direct consequence of soundness of CCBS and soundness of the encoding (Proposition \ref{prop-valid}).

\begin{proposition}
The SMT-CBS$^\mathcal{R}$ algorithm returns makespan optimal solution for any solvable MAPF$^\mathcal{R}$ instance $\Sigma^\mathcal{R}$.
\label{prop-SMT}
\end{proposition}



 \begin{algorithm}[t]
\begin{footnotesize}
\SetKwBlock{NRICL}{SMT-CBS-Fixed$^\mathcal{R}$($\Sigma^\mathcal{R}$, $cons$, $\mu$)}{end} \NRICL{
	    $\textsc{Rdd} \gets$ build-RDDs($\Sigma^\mathcal{R}$, $cons$, $\mu$)\\
	    $\mathcal{F}(\mu) \gets$ encode-Basic$(\textsc{Rdd},\Sigma^\mathcal{R},cons,\mu)$\\
	    \While {$\mathit{TRUE}$}{
	        $assignment \gets$ consult-SAT-Solver$(\mathcal{F}(\mu))$\\
	        \If {$assignment \neq UNSAT$}{
	            $\pi \gets$ extract-Solution$(assignment)$\\
	            $collisions \gets$ validate-Plans($\pi$)\\
                   \If {$collisions = \emptyset$}{
                      \Return $(\pi, \emptyset, \mathit{UNDEF})$\\
                   }
                   \For{each $(m_i \times m_j) \in collisions$ where $m_i = (a_i,(u_i,v_i),[t_i^0,t_i^+))$ and $m_j=(a_j,(u_j,v_j),[t_j^0,t_j^+))$}{
                      $\mathcal{F}(\mu)$$\gets$$\mathcal{F}(\mu)$$\wedge$$(\neg \mathcal{E}_{u_i,v_i}^{t_i^0,t_i^+}(a_i)$$\vee$$\neg \mathcal{E}_{u_j,v_j}^{t_j^0,t_j^+}(a_j))$\\
                      $([\tau_i^0, \tau_i^+);[\tau_j^0, \tau_j^+)) \gets$ resolve-Collision($m_i$,$m_j$)\\                   
                      $cons \gets cons \cup \{[(a_i,(u_i,v_i),[\tau_i^0,\tau_i^+));$ $(a_j,(u_j,v_j),[\tau_j^0,\tau_j^+))]\}$\\                   
                   }                  
	    		$\textsc{Rdd} \gets$build-RDDs($\Sigma^\mathcal{R}$, $cons$, $\mu$)\\
	    		$\mathcal{F}(\mu) \gets$ augment-Basic$(\textsc{Rdd},\Sigma^\mathcal{R},cons)$\\                   
               }
         }
	   $\mu_{next} \gets$ min$\{ t \;|\; (u,t) \in X_i \wedge t > \mu$\\
	   $\;\;$ where $\textsc{Rdd}_i=(X_i,E_i)$ for $i=1,2,...,k)\}$\\
         \Return {(UNSAT, $cons$, $\mu_{next}$)}\\
}
\caption{Low-level of SMT-CBS$^\mathcal{R}$} \label{alg-SMTCBS-low}
\end{footnotesize}
\end{algorithm}

\section{Experimental Evaluation}

We implemented SMT-CBS$^\mathcal{R}$ in C++ to evaluate its performance and compared it with a version of CCBS adapted for the makespan objective \footnote{To enable reproducibility of presented results we provide complete source code of our solvers on \texttt{https://github.com/surynek/boOX}.} 


SMT-CBS$^\mathcal{R}$ was implemented on top of Glucose 4 SAT solver \cite{DBLP:conf/ijcai/AudemardS09} which ranks among the best SAT solvers according to recent SAT solver competitions \cite{DBLP:conf/aaai/BalyoHJ17}. Whenever possible the SAT solver was consulted in the incremental mode.

The actual implementation builds RDDs in a more sophisticated way than presented pedagogically in Algorithm \ref{alg-RDD-build}. The implementation prunes out decisions from that the goal vertex cannot be reached under given makespan bound $\mu_{max}$: whenever we have a decision $(u,t)$ such that $t + \Delta t > \mu_{max}$, where  $\Delta t = dist_{estimate}(u, \alpha_+(a)) / v_a $ and $dist_{estimate}$ is a lower bound estimate of the distance between a pair of vertices, we rule out that decision from further consideration.

In case of CCBS, we used the existing C++ implementation for the sum-of-costs objective \cite{DBLP:conf/ijcai/AndreychukYAS19} and modified it for makespan while we tried to preserve its heuristics from the sum-of-costs case. 

\subsection{Benchmarks and Setup}
SMT-CBS$^\mathcal{R}$ and CCBS were tested on benchmarks from the $\mathtt{movinai.com}$ collection \cite{sturtevant2012benchmarks}. We tested algorithms on three categories of benchmarks: 

\begin{enumerate}[label=(\roman*)]
    \item {\bf small} empty grids (presented representative benchmark $\mathtt{empty}$\nobreakdash-$\mathtt{16}$\nobreakdash-$\mathtt{16}$),
    \item {\bf medium} sized grids with regular obstacles (presented $\mathtt{maze}$\nobreakdash-$\mathtt{32}$\nobreakdash-$\mathtt{32}$\nobreakdash-$\mathtt{4}$),
    \item {\bf large} game maps (presented $\mathtt{ost003d}$).
\end{enumerate}

In each benchmark, we interconnected cells using the $2^K$-neighborhood \cite{DBLP:conf/aaai/RiveraHB17} for $K=3,4,5$ - the same style of generating benchmarks as used in \cite{DBLP:conf/ijcai/AndreychukYAS19} ($K=2$ corresponds to MAPF hence not omitted). Instances consisting of $k$ agents were generated by taking first $k$ agents from random scenario files accompanying each benchmark on $\mathtt{movinai.com}$. Having 25 scenarios for each benchmarks this yields to 25 instances per number of agents.

\begin{figure}[h]
    \centering
    \vspace{-0.1cm}
    \includegraphics[trim={2.7cm 21cm 2.4cm 2.6cm},clip,width=0.75\textwidth]{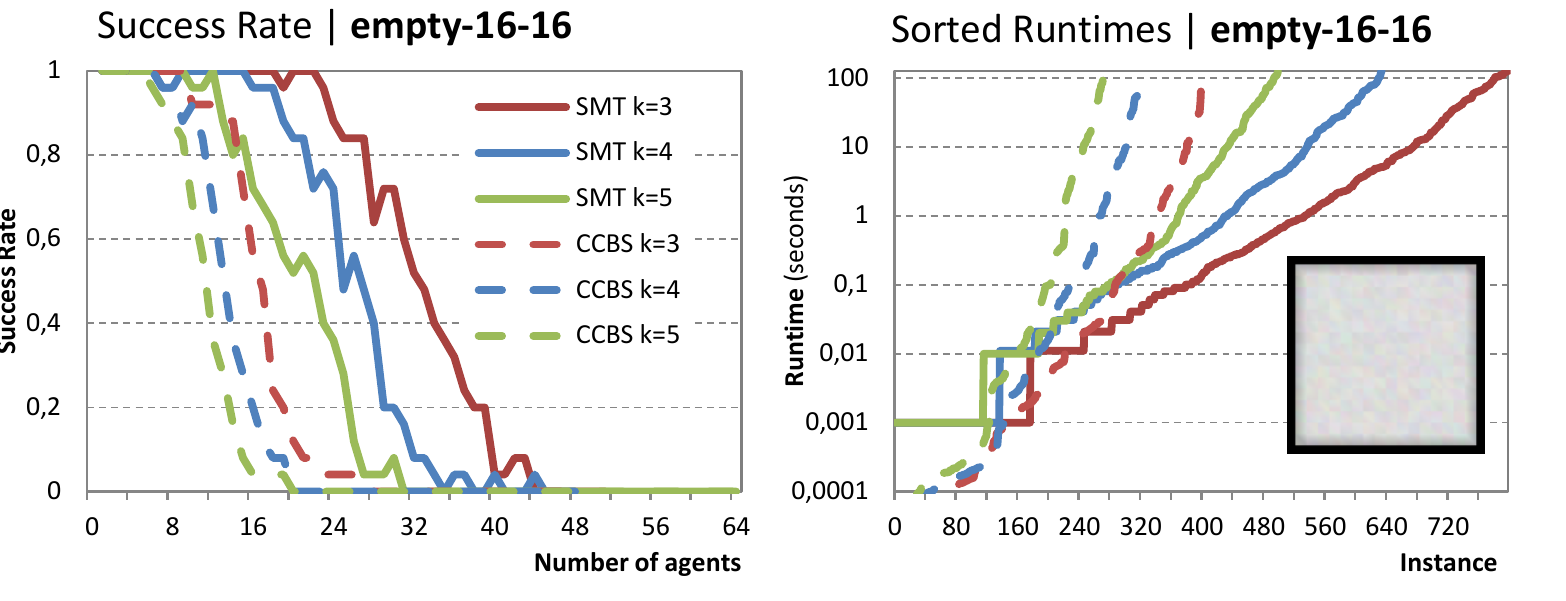}
    \vspace{-0.6cm}
    \caption{Comparison of SMT-CBS$^\mathcal{R}$ and CCBS on $\mathtt{empty}$-$\mathtt{16}$-$\mathtt{16}$.}
    \label{expr_empty}
\end{figure}

\begin{figure}[h]
    \centering
    \vspace{-0.1cm}
    \includegraphics[trim={2.7cm 21cm 2.4cm 2.6cm},clip,width=0.75\textwidth]{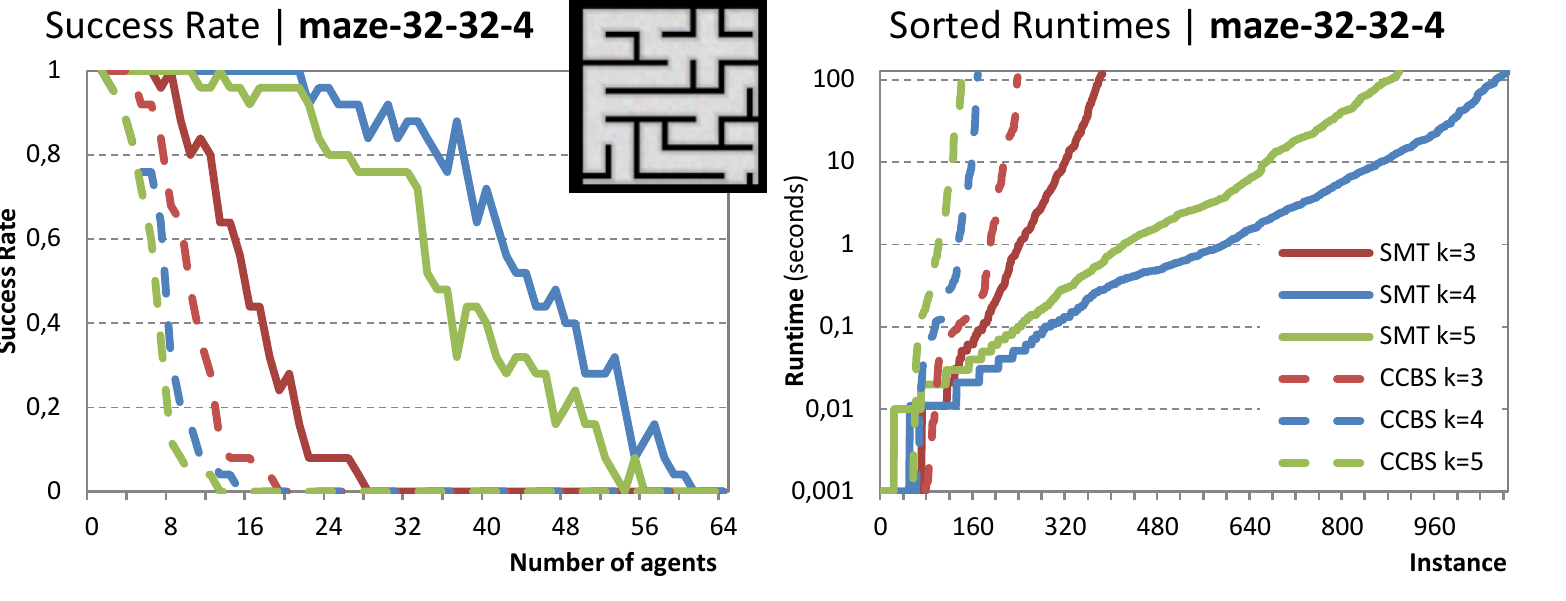}
    \vspace{-0.6cm}
    \caption{Comparison of SMT-CBS$^\mathcal{R}$ and CCBS on $\mathtt{maze}$-$\mathtt{32}$-$\mathtt{32}$-$\mathtt{4}$.}
    \label{expr_maze}
\end{figure}

\begin{figure}[h]
    \centering
    \vspace{-0.1cm}
    \includegraphics[trim={2.7cm 21cm 2.4cm 2.6cm},clip,width=0.75\textwidth]{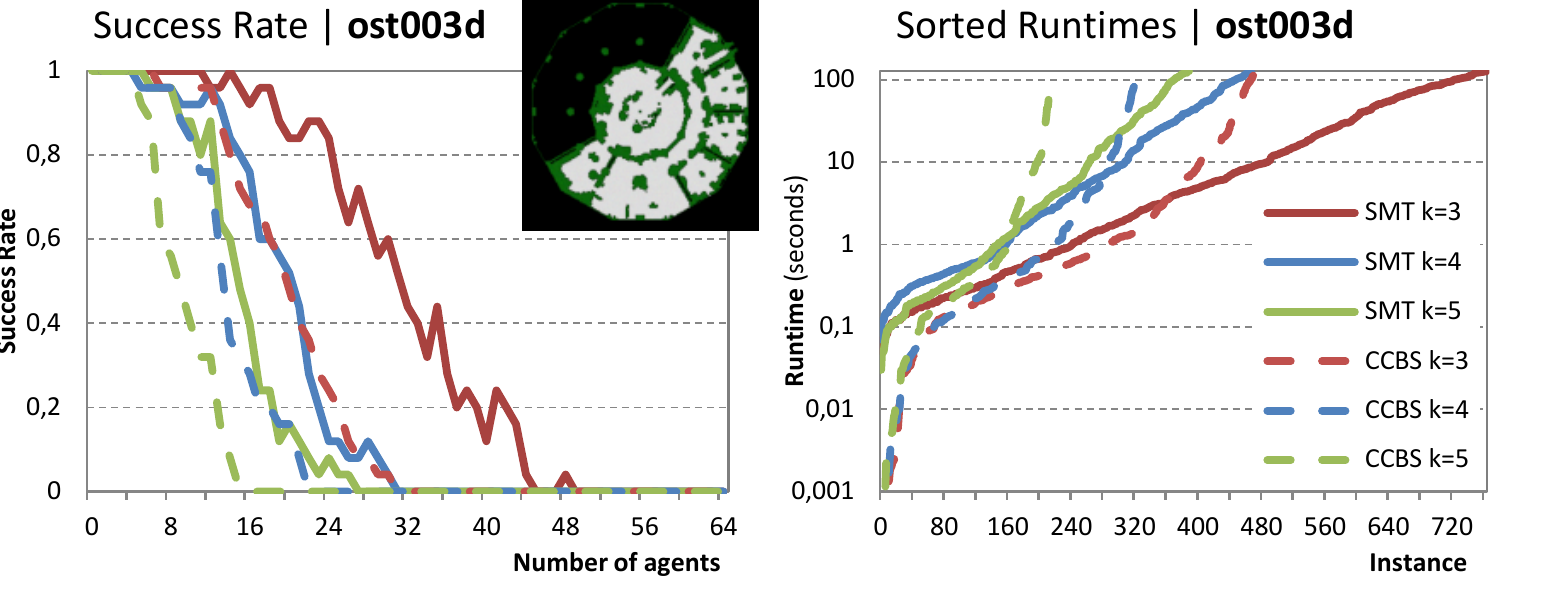}
    \vspace{-0.6cm}
    \caption{Comparison of SMT-CBS$^\mathcal{R}$ and CCBS on $\mathtt{ost003d}$.}
    \label{expr_ost003d}
\end{figure}

Part of the results obtained in our experimentation is presented in this section\footnote {All experiments were run on a system with Ryzen 7 3.0 GHz, 16 GB RAM, under Ubuntu Linux 18.}. For each presented benchmark we show {\em success rate} as a function of the number of agents. That is, we calculate the ratio out of 25 instances per number of agents where the tested algorithm finished under the timeout of 120 seconds. In addition to this, we also show concrete runtimes sorted in the ascending order. Results for one selected representative benchmark from each category are shown in Figures \ref{expr_empty}, \ref{expr_maze}, and \ref{expr_ost003d}.

The observable trend is that the difficulty of the problem increases with increasing size of the $K-$neighborhood with notable exception of $\mathtt{maze}$\nobreakdash-$\mathtt{32}$\nobreakdash-$\mathtt{32}$\nobreakdash-$\mathtt{4}$ for $K=4$ and $K=5$ which turned out to be easier than $K=3$ for SMT-CBS$^\mathcal{R}$.

Throughout all benchmarks SMT-CBS$^\mathcal{R}$ tends to outperform CCBS. The dominance of SMT-CBS$^\mathcal{R}$ is most visible in medium sized benchmarks. CCBS is, on the other hand, faster in instances containing few agents. The gap between SMT-CBS$^\mathcal{R}$ and CCBS is smallest in large maps where SMT-CBS$^\mathcal{R}$ struggles with relatively big overhead caused by the big size of the map (the encoding is proportionally big). Here SMT-CBS$^\mathcal{R}$ wins only in hard cases.

\section{Discussion and Conclusion}

We suggested a novel algorithm for the makespan optimal solving of the multi-agent path finding problem in continuous time and space called SMT-CBS$^\mathcal{R}$ based on {\em satisfiability modulo theories} (SMT). Our approach builds on the idea of treating constraints lazily as suggested in the CBS algorithm but instead of branching the search after encountering a conflict we refine the propositional model with the conflict elimination disjunctive constraint as it has been done in previous application of SMT in the standard MAPF.

The major obstacle in using SMT and propositional reasoning not faced previously with the standard MAPF is that decision variables cannot be determined in advance straightforwardly in the continuous case. We hence suggested a novel decision variable generation approach that enumerates new decisions after discovering new collisions.

We compared SMT-CBS$^\mathcal{R}$ with CCBS \cite{DBLP:conf/ijcai/AndreychukYAS19}, currently the only alternative algorithm for MAPF$^\mathcal{R}$ that modifies the standard CBS algorithm, on a number of benchmarks. The outcome of our comparison is that SMT-CBS$^\mathcal{R}$ performs well against CCBS. The best results SMT-CBS$^\mathcal{R}$ are observable on medium sized benchmarks with regular obstacles. We attribute the better runtime results of SMT-CBS$^\mathcal{R}$ to more efficient handling of disjunctive conflicts in the underlying SAT solver through {\em propagation}, {\em clause learning}, and other mechanisms. On the other hand SMT-CBS$^\mathcal{R}$ is less efficient on large instances with few agents.


For the future work we assume extending the concept of SMT-based approach for MAPF$^\mathcal{R}$ with other cumulative cost functions other than the makespan such as the {\em sum-of-costs} \cite{DBLP:journals/ai/SharonSGF13,DBLP:conf/aips/FelnerLB00KK18}. We also plan to extend the RDD generation scheme to directional agents where we need to add the third dimension in addition to space (vertices) and time: {\em direction} (angle). The work on MAPF$^\mathcal{R}$ could be further developed into multi-robot motion planning in continuous configuration spaces \cite{DBLP:books/daglib/0016830}.


\bibliographystyle{splncs04}
\bibliography{bibfile}

\end{document}